\documentclass[letterpaper, 10 pt, conference]{ieeeconf}  

\IEEEoverridecommandlockouts                              

\usepackage{amsbsy}
\usepackage{multirow}
\usepackage{multicol}
\usepackage{graphicx}
\usepackage{url}
\usepackage{comment}
\usepackage{xcolor}
\usepackage{amsmath}
\usepackage{amssymb}
\usepackage{amsfonts}
\usepackage{bbm}
\usepackage{dsfont}

\newcommand{\norm}[1]{\left\lVert#1\right\rVert}

\title{\LARGE \bf
MEMROC: Multi-Eye to Mobile RObot Calibration}

\author{Davide Allegro$^{1 }$, Matteo Terreran$^{1\, \dagger}$ and Stefano Ghidoni$^{1}$
\thanks{$\dagger$ Corresponding Author {\tt\small matteo.terreran@dei.unipd.it}}%
\thanks{$^{1}$Intelligent Autonomous System Lab, Department of Information Engineering, University of Padova, Padua, Italy.
        {\tt\small \{allegrodav, terreran, ghidoni\}@dei.unipd.it}}
}

\begin{document}

\maketitle
\thispagestyle{empty}
\pagestyle{empty}

\begin{abstract}
This paper presents MEMROC (Multi-Eye to Mobile RObot Calibration), a novel motion-based calibration method that simplifies the process of accurately calibrating multiple cameras relative to a mobile robot's reference frame. 
MEMROC utilizes a known calibration pattern to facilitate accurate calibration with a lower number of images during the optimization process. Additionally, it leverages robust ground plane detection for comprehensive 6-DoF extrinsic calibration, overcoming a critical limitation of many existing methods that struggle to estimate the complete camera pose.
The proposed method addresses the need for frequent recalibration in dynamic environments, where cameras may shift slightly or alter their positions due to daily usage, operational adjustments, or vibrations from mobile robot movements. MEMROC exhibits remarkable robustness to noisy odometry data, requiring minimal calibration input data. This combination makes it highly suitable for daily operations involving mobile robots. A comprehensive set of experiments on both synthetic and real data proves MEMROC's efficiency, surpassing existing state-of-the-art methods in terms of accuracy, robustness, and ease of use. To facilitate further research, we have made our code publicly available\footnote{\url{https://github.com/davidea97/MEMROC.git}}.
\end{abstract}

\section{INTRODUCTION}

\begin{figure}[ht!]
  \centering
  \includegraphics[width=1\linewidth]{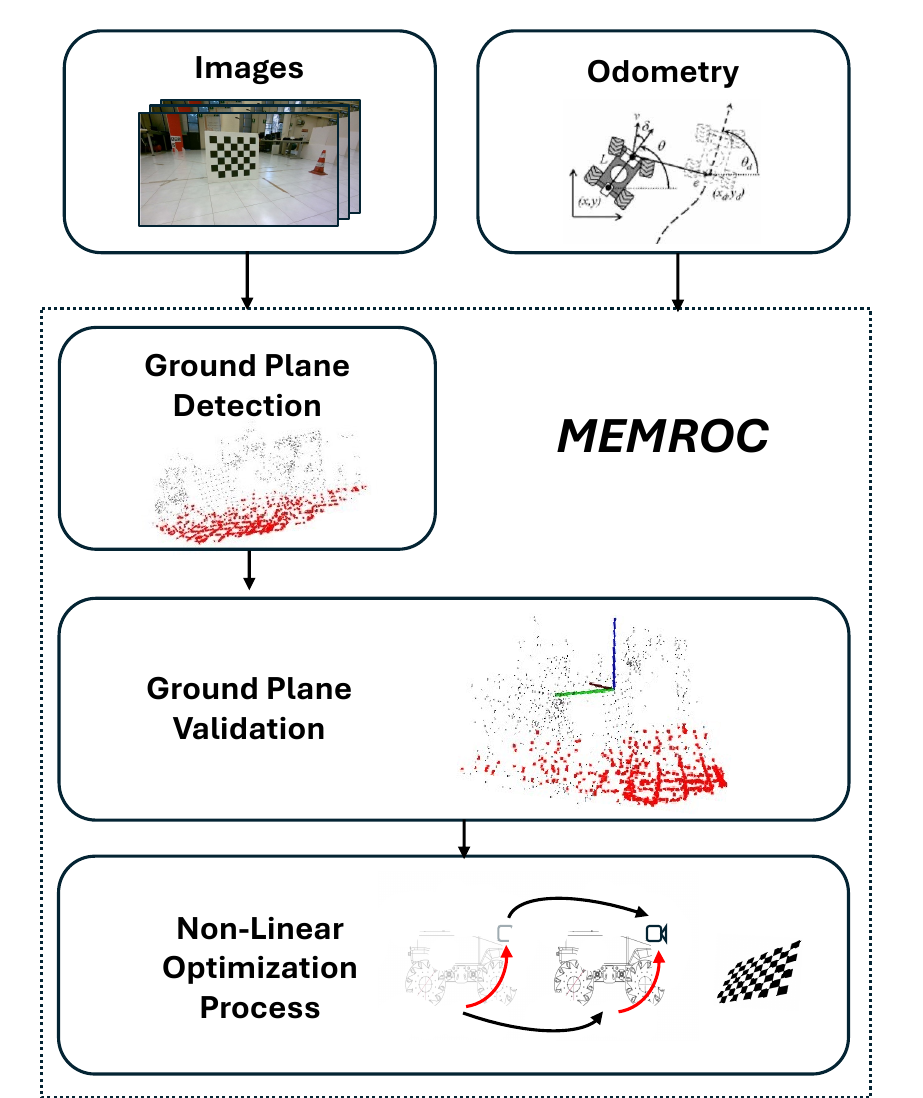}
  \caption{MEMROC Overview. The system is structured into three primary components: the Ground Plane Detection module, responsible for identifying the ground plane; the Ground Plane Validation module, which ensures the ground plane's alignment with the mobile robot; and the Motion-Based Calibration Process, the concluding stage that calibrates the system.}
  \label{fig:block}
\end{figure}

Mobile robots rely on one or more onboard sensors to perceive the environment and determine their location. Visual cameras are a popular choice due to their affordability and ability to capture extensive information from the scene: this led to a widespread use of such sensors in robotic navigation~\cite{yuan2019multisensor,chiang2019seamless} and autonomous driving~\cite{henein2020dynamic,fernandes2021point}.
When the robot moves, onboard cameras frame the environment from multiple points of view, that can be related by means of pure vision algorithms or considering the robot odometry.
Accurate alignment of visual data with respect to the robot reference frame is crucial for safe navigation~\cite{fusaro2022pushing} and for enhancing perception, localization, and the robot's ability to interact effectively with its environment~\cite{zheng2019visual,evangelidis2021revisiting}. To ensure a robot effectively utilizes visual information of its surroundings, it is required that the data collected by its cameras are expressed with respect to the robot reference frame—typically situated at the central point among its four wheels on the ground.
This is why accurately calibrating the various cameras with respect to the mobile robot is essential.

Traditional calibration methods rely on specific infrastructures for precise sensor installation~\cite{koppanyi2018experiences,horn2023extrinsic}, the need for environments equipped with numerous markers~\cite{xie2018infrastructure,antonelli2010simultaneous}, the requirement for supplementary sensors such as lidars to determine feature point correspondences among sensors~\cite{gao2010line}, or the necessity of manual image collection processes~\cite{wang2022accurate}. These approaches are not suitable for frequent use, especially in dynamic real-world environment, which can cause minor camera displacements due to everyday operation, adjustments, or vibrations from robot movement.  Frequent recalibration is necessary in such scenarios, and existing methods are often cumbersome and time-consuming. Furthermore, many approaches focus on calibrating cameras relative to each other requiring overlapping fields of view and neglecting the fundamental objective of expressing sensor data with the robot's reference frame~\cite{yan2022opencalib,furgale2013unified,jiao2019automatic}. Motion-based calibration methods tackle these limitations by exploiting robot motion for calibrating cameras with respect to the robot. However, a key limitation arises from the inherent planar constraints of most mobile robots, preventing them from estimating the full 6-DoF of the camera's extrinsic parameters solely through incremental movements~\cite{della2019unified,guo2012analytical}.
To overcome such limitation, Zuñiga's method~\cite{zuniga2019automatic} estimates all the 6-DoF through the ground plane detection, but it requires a large amount of data and a refinement procedure to be accurate; it also exhibits high sensitivity to noisy odometry data---a common issue due to odometry drift that limits its robustness~\cite{zuniga2019automatic}. Learning-based calibration methods are faster but might offer a limited accuracy and adaptability to changing environments, as they often require adherence to specific vehicle trajectories or the presence of well-defined road features for successful calibration, hindering their real-world applicability~\cite{meyer2021automatic}. 
%
The limitations of existing methods highlight the need for faster, easier, and more robust sensor calibration techniques that consider the robot's reference frame. This would enable robots to maintain optimal performance through frequent recalibration and ultimately enhance their ability to interact effectively with their surroundings.

This paper presents MEMROC: Multi-Eye to Mobile RObot Calibration.
This is a novel motion-based calibration method for mobile robots which efficiently determines the full 6-DoF calibration parameters of multiple cameras relative to the robot incorporating a robust mechanism for the ground plane detection and its validation. 
Notably, the overlap of camera fields of view is \emph{not} a prerequisite for this method; however, the optimization process is capable of taking advantage of the overlap, if available, to refine the geometric constraints between cameras, ensuring enhanced accuracy. 
After comprehensive experiments on both synthetic and real-world data, MEMROC demonstrated exceptional accuracy while requiring minimal data inputs. This was achieved through the observation of a calibration pattern, such as a checkerboard, from fewer than 20 viewpoints, and capturing the ground plane at least once. Unlike conventional approaches, MEMROC's non-linear optimization technique shows remarkable robustness against noisy odometry data.
This capability greatly increases the usability of autonomous mobile robots in daily environments. 

Summarizing, this paper presents the following main contributions:
\begin{enumerate}
    \item The concept of MEMROC, a novel motion-based calibration method  for the estimation of the full 6-DoF pose of multiple cameras with respect to the robot;
    \item A comprehensive performance evaluation of the proposed method by means of simulations and real-world experiments in both indoor and outdoor settings, demonstrating its robustness and accuracy;
    \item A thorough comparative analysis focusing on how the precision of MEMROC is minimally dependent on the number of images used for calibration, showing its superior precision with minimal data requirements against other state-of-the-art methods;
    \item An extensive, publicly available dataset, including 3000 synthetic and 3000 real images collected with a mobile robot, intended to facilitate the evaluation of future camera to robot calibration methods.
\end{enumerate}

The remainder of the paper is organized as follows: Section~\ref{sec:related_works} reviews state-of-the-art methods on camera calibration for mobile robots. Section~\ref{sec:method} delves into the theoretical foundation of our method, while Section~\ref{sec:dataset} outlines the dataset collection. Section~\ref{sec:experiments} provides a detailed analysis of results on the collected dataset against state-of-the-art methods, in particular Sections~\ref{subsec:synthetic} and \ref{subsec:real} report outcomes in synthetic and real-world scenarios, respectively. Finally, Section~\ref{sec:conclusions} concludes and outlines future research directions.
\begin{figure*}[t]
  \centering
  \includegraphics[width=0.75\linewidth]{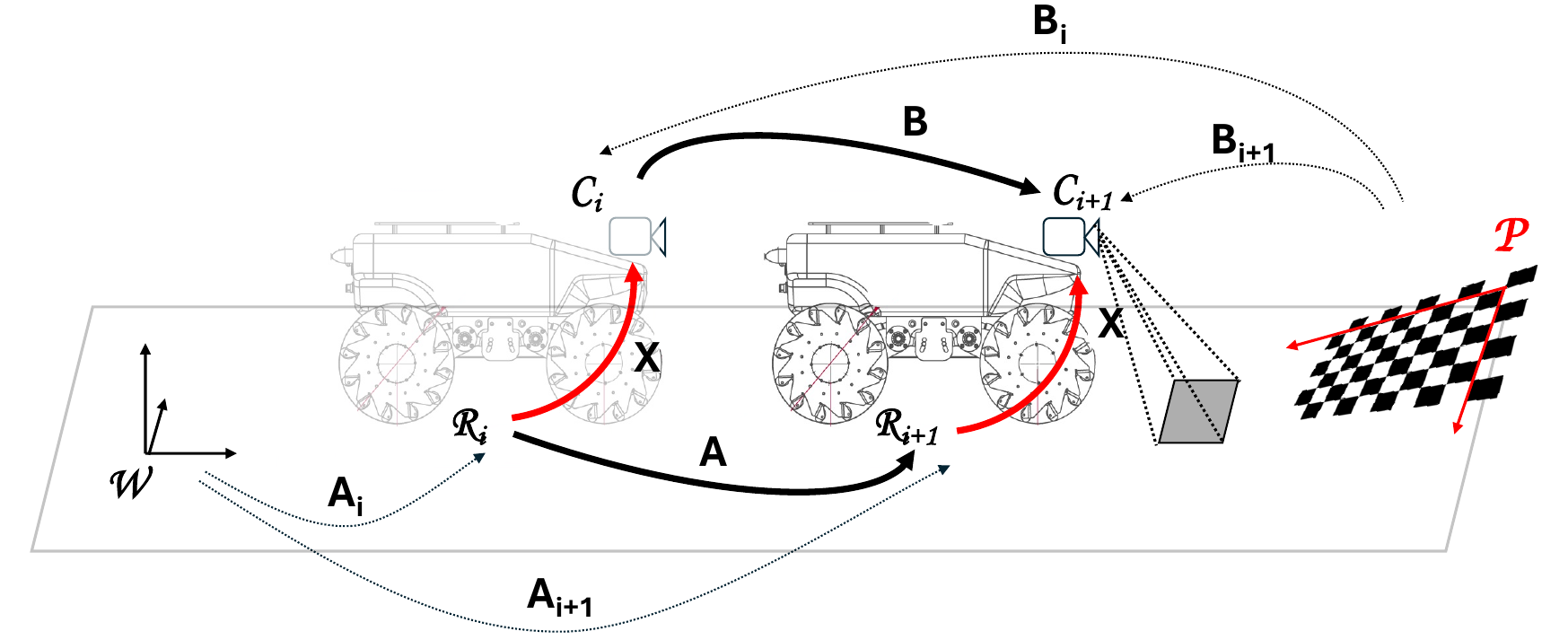}
  \caption{Formulation of the motion-based calibration method. A mobile robot $R$ equipped with a camera $C$ moves from pose $i$ to $i+1$ towards a calibration pattern $P$ in the scene. $A_i$ represents the odometry of the mobile robot with respect to the fixed frame $W$ (i.e., the robot starting position), while $B_i$ represents the camera pose with respect to the calibration pattern. X denotes the camera to robot transformation to be estimated.}
  \label{fig:method}
\end{figure*}
\section{RELATED WORKS}
\label{sec:related_works}
Multi-sensor calibration on mobile robots can be categorized into three main approaches: appearance-based, motion-based, and learning-based methods.
Appearance-based methods leverage prior knowledge about the environment. For instance, Gómez et al.~\cite{gomez2015extrinsic} employed scene corners (orthogonal trihedrons) to calibrate a 2D laser scanner and a camera. Similarly, Gao et al.~\cite{gao2010line} proposed a multi-LiDAR calibration method using point constraints from retro-reflective targets placed in the scene. Additionally, Choi et al.~\cite{choi2015extrinsic} utilized the appearance of two orthogonal planes to determine the spatial offset between dual 2D LiDARs.
However, appearance-based calibration methods are typically designed for calibrating specific sensor pairs, and applying them to a multi-sensor system with various combinations can become complex and impractical, as noted by Xie et al.~\cite{xie2018infrastructure}.
%
More recently, Meyer et al.~\cite{meyer2021automatic} proposed a learning-based method that leverages road features like lane markings or edges to calibrate yaw, and ground surface orientation for pitch and roll. However, this approach requires the vehicle to travel along a straight path with well-defined road markings. In contrast, Yan et al.~\cite{yan2023sensorx2car} introduced SensorX2Car, which utilizes vanishing points and the horizon line to estimate the 3-DoF rotation between the sensor and the robot. This method, however, relies on the assumption that the vehicle's orientation always aligns with its trajectory, which is not true in the case of cameras mounted on the side of the vehicles~\cite{wang2022accurate}.


Motion-based approaches draw inspiration from the well-established hand-eye calibration problem~\cite{horaud1995hand}. They estimate the extrinsic parameters (relative poses) between sensors by analyzing their combined motions. This typically involves solving an equation of the form $AX = XB$, where A and B represent the motion of two sensors installed on the same robot and X represents the unknown transformation.
Censi et al.~\cite{censi2013simultaneous} leverage this approach by combining wheel odometry and exteroceptive sensor data for multi-sensor calibration on mobile robots. Similarly, Kummerle et al.~\cite{kummerle2011simultaneous} propose an algorithm for visual-odometry calibration using motion data. Della Corte et al.~\cite{della2019unified} go a step further by estimating sensor time delays alongside extrinsic parameters within their motion-based calibration framework. 
However, a limitation of these methods is their inability to estimate all 6-DoF describing the sensor-to-robot transformations due to the unobservability of the z-coordinate~\cite{guo2012analytical}. Works like Heng et al.'s CamOdoCal~\cite{heng2013camodocal} address this problem by estimating full 6-DoF for multiple cameras for identifying feature point correspondences between the current frame of each camera and the frames from every other camera's historical data, resulting in extensive and time-consuming calibration processes. Zuñiga's RobotAutocalib~\cite{zuniga2019automatic} estimates full 6-DoF through ground plane detection, but suffers from inconsistencies due to its two-step approach (separately x, y, yaw and then the remaining 3-DoF).
Furthermore, it needs a large number of images (often exceeding 50) to accurately estimate the ground plane using Structure-from-Motion (SfM) pipelines. Additionally, it only considers constraints between individual sensors and a reference, neglecting consistency across different sensor transformations.



\section{METHOD}
\label{sec:method}

In this section, our MEMROC approach for full 6-DoF camera to robot calibration is presented and described. Our method follows a motion-based approach where calibration is obtained by minimizing the general $\|AX-XB\|^{2}$ cost function, introducing two main elements: a known calibration pattern fixed in the scene and a ground plane estimation module.
The use of a known pattern (e.g. checkerboard) allows to overcome the inherent scale ambiguity in motion-based approaches, leading to several advantages: (i) more accurate camera motion estimation and (ii) a lower number of images needed in the optimization process compared to methods relying solely on visual features.
Ground plane estimation, on the other hand, allows to address z-coordinate unobservability and achieve full 6-DoF calibration. It provides a set of measurements of the camera's height relative to the ground which are then used as constraints in the optimization process.

A schematic overview of MEMROC is shown in Figure~\ref{fig:block}, representing its main modules. First, a set of pairs of robot poses and images is collected by moving the robot in front of a calibration pattern as depicted in Figure~\ref{fig:method}; robot poses $A_i$ are obtained from odometry, while images allow to estimate camera poses $B_i$ with respect to the calibration pattern by means of PnP (Perspective-n-Point) algorithm. Images are also used to compute 3D correspondences based on feature matching for ground plane detection and validation based on RANSAC and Singular Value Decomposition.
Finally, robot and camera poses are used in the motion-based optimization process with ground plane measurements as constraint.


\subsection{Ground plane detection}
\label{subsec:ground}

The general idea is to estimate the ground plane equation from camera measurements, in order to compute its height from the ground and use such information in the optimization process. 
In particular, assuming a 3D representation of the scene in the camera frame is available (e.g., pointcloud), we identify the largest plane in such representation using the RANSAC (RANdom SAmple Consensus) method~\cite{li2017improved}. 

When considering RGB-D cameras, the 3D representation of the scene can be easily obtained from the pointcloud provided by the camera. However, the quality of the pointcloud can be very inaccurate in outdoor scenarios. Therefore, we consider an alternative approach to obtain 3D points of the scene based on feature matching between consecutive frames.
%
%
In particular, given two consecutive images $I_i$ and $I_{i+1}$, for each one of them a set of keypoints and descriptors is extracted (e.g., SIFT or SuperPoint~\cite{detone2018superpoint}). 
Let $\mathcal{M} = \{(p_{1,i}, p_{1,i+1}),\dots, (p_{M,i}, p_{M,i+1})\}$ the set of matches between the two images computed by a feature matcher (e.g., LightGlue~\cite{lindenberger2023lightglue}). 
Based on the matches $\mathcal{M}$ and the absolute camera poses $B_i$, $B_{i+1}$ provided by the PnP algorithm, it is possible to compute the 3D coordinates of the matches by means of stereo triangulation using direct linear transform (DLT)~\cite{Hartley2004}. 
Note that by using the absolute camera poses, the final result is a set of 3D points expressed with respect to the pattern reference frame $\mathcal{P}$, as depicted in Figure~\ref{fig:block}. Such points constitutes the 3D scene representation used for estimating the ground plane with RANSAC (red points in Figure~\ref{fig:block}), which provides the equation of the largest plane in the scene expressed in the general equation form
\begin{equation}
    ax + by + cz + d = 0
\end{equation}
where $a$, $b$ and $c$ are the components of the normal vector $\vec{n} = (a,b,c)$ perpendicular to the plane and $d$ is the minimum distance from the plane, which also represent the height of the camera from the ground plane to be used in the constrained calibration process.

\subsection{Ground plane validation}
\label{subsec:refinement}

The ground plane detection algorithm described above aims to compute the plane equation of the largest plane in the reconstructed scene. However, such plane could it be different from the ground plane, especially in crowded indoor scenarios (e.g., vertical walls, large flat objects) so a validation procedure to filter out wrong detections is needed. 

When moving the robot, since the camera is rigidly attached to the robot, the robot and the camera perform the same trajectory but expressed in two different fixed reference frames (i.e., world and pattern respectively).
Let $a = \{a_1, a_2, \dots, a_n\}$ and $b = \{b_1, b_2, \dots, b_n\}$ two sets of corresponding 3D points, obtained from the translation part of matrices $A_i$ and $B_i$ respectively. By means of Singular Value Decomposition (SVD)~\cite{sorkine2017least}, it is possible to find a rigid transformation that optimally aligns the two sets in the least squares sense:
\begin{equation}
    (R,t) = \operatorname*{argmin}_{R \in SO(3), t\in \mathbb{R}^3} \sum_{i=1}^n || (Rb_i + t) - a_i ||^2
\end{equation}
%
%
The obtained rigid transformation describes the roto-translation  $(R^W_P,t^W_P)$ between the world and the pattern reference frame. 
Although the estimated translation is affected by a bias term (it assumes that the 3D points are coincident, while $a$ and $b$ are only parallel planes which differ for an offset term), the estimated rotation can be used to transform the 3D points of the scene obtained Section~\ref{subsec:ground} in the world reference frame, $p^W = R^W_P \, p^P$. 
By introducing such operation before the RANSAC-based plane detection, the estimated plane equation is then expressed in world coordinates and can be easily verify the correct alignment of the normal of the plane with the z-axis of the world reference frame. If the estimated plane is not parallel to the ground, it is rejected.



\subsection{Camera to robot calibration}
\label{subsec:cametatorobot}
%
%
As illustrated in Figure~\ref{fig:method}, for each robot movement, we collect transformations $A_i$ and $B_i$ representing the robot's absolute pose at time step $i$ relative to its initial position (i.e., world frame W), and the pose of the camera with respect to the fixed calibration pattern reference frame.
%
%
However, robot poses obtained by odometry are generally noisy and affected by drift errors. This makes such data unsuitable for the calibration process, making it preferable to consider relative movements of the robot to limit the effect of drift in odometry. 
Therefore, in MEMROC, we aim to minimizes the cost function defined as 
\begin{equation}
    c = \sum_{i=0}^{N-1} \left\|AX-XB\right\|^2,
    \label{eq:general_cost}
\end{equation}
where $A$ represents the robot's incremental transformation within a 2D plane, $B$ denotes the camera's incremental motion, and $X$ is the camera to robot transformation matrix we aim to determine through the calibration process. 
%
Given $N$ samples of incremental motion, we define our calibration methodology through the objective of minimizing a specifically formulated cost function, detailed in equation (\ref{eq:minimization_eq}). 


\begin{gather}
c_{Zi} = \chi_{i}\norm{[T^{R}_{C}]_{z}-\overline{z_{i}}}^{2}  \\
c = \sum_{i=0}^{N-1} \left( \norm{T_{R_{i+1}}^{R_{i}}T^{R}_{C} - T^{R}_{C}T_{C_{i+1}}^{C_{i}}}^{2} + c_{Zi} \right)
\label{eq:minimization_eq}
\end{gather}

In this formulation, $T_{R_{i+1}}^{R_{i}}$ and $T_{C_{i+1}}^{C_{i}}$ symbolize the incremental motions of the robot and the camera, respectively, capturing their transition from one state to the next. Our objective is to accurately determine the transformation $T^{R}_{C}$, representing the rototranslation between the camera and the robot.
An additional term $c_{Zi}$ in the cost function aims to reduce the difference between the z-axis coordinate of the desired transformation and the camera's elevation, $\overline{z_{i}}$, as identified through ground plane detection (i.e., $\overline{z_{i}}=d$). The variable $\chi_{i}$, which can either be 0 or 1, signifies whether the $i^{th}$ pose yields a valid measure of the camera's height above the ground plane.
This approach allows for the precise estimation of the camera's 6-DoF pose relative to the robot, moving beyond simple ground plane detection to integrate a comprehensive set of spatial data into the calibration effort. Remarkably, a single $z_{i}$ measurement suffices for this calibration. Moreover, this optimization process calibrates each camera independently with respect to the robot, thereby eliminating the necessity for overlapping fields of view among sensors. This strategy not only simplifies the calibration process but also expands its versatility to support various sensor arrangements without sacrificing precision.



\subsection{Multi-sensor joint optimization}
\label{subsec:multicametatorobot}
%
%
When multiple cameras are mounted on the vehicle, each camera can be calibrated independently by means of the optimization process described in Section~\ref{subsec:cametatorobot}. However, in the case that the cameras have overlapping fields of view, they can simultaneously detect the calibration pattern providing additional information on the relative pose between them.
In such a scenario, it is possible to include mutual relations between cameras in the cost function to optimize not only the extrinsic parameters of each camera relative to the robot frame, but also the relative poses between pairs of cameras.

Consider the scheme in Figure~\ref{fig:method} and assume that two cameras $C_1$ and $C_2$ are mounted on the vehicle in the front direction. When the robot moves from pose $i$ to pose $i+1$, if both cameras can detect the calibration pattern, the relation between their incremental motion can be expressed as:
\begin{equation}
     T_{C_{1,{i+1}}}^{C_{1,{i}}}T^{C_{1}}_{R}T^{R}_{C_{2}} = T^{C_{1}}_{R}T^{R}_{C_{2}}T_{C_{2,{i+1}}}^{C_{2,{i}}}
\end{equation}
%
which can be easily reformulated in the general $AX = XB$ equation by assuming $X = T^{C_{1}}_{R} T^{R}_{C_{2}} = (T^{R}_{C_{1}})^{-1} T^{R}_{C_{2}}$. 
Note that such unknown matrix can be expressed as the product of the two camera to robot transformations $T^{R}_{C_{1}}$ and $T^{R}_{C_{2}}$ to be estimated in the calibration process, therefore providing an additional constraint to be exploited in the optimization.   

Let denote by $\mathcal{S}_i$ the set of camera pairs $(j,k)$ which detects the calibration pattern at step $i$, and $\mathds{1}_{\mathcal{S}_i}$ the corresponding indicator function. With such notation, the constraint describing the joint detection by multiple $M$ cameras at step $i$ can be expressed as:
\begin{equation}
c_{Ji} = \sum_{j=1}^{M}\sum_{k = j}^{M}\mathds{1}_{\mathcal{S}_i} \norm{T_{C_{j,i+1}}^{C_{j,i}}T^{C_{j}}_{R}T^{R}_{C_{k}} - T^{C_{j}}_{R}T^{R}_{C_{k}}T_{C_{k,i+1}}^{C_{k,i}} }^{2}
\label{eq:first_joint_minimization_eq}
\end{equation}

Finally, including the joint constraint ~(\ref{eq:first_joint_minimization_eq}) in the cost function~(\ref{eq:minimization_eq}), the overall cost function can be expressed as:
\begin{equation}
\begin{split}
  c = &\sum_{i=0}^{N-1} \left( \sum_{j=1}^{M} 
  \left( 
  \norm{T_{R_{i+1}}^{R_{i}}T^{R}_{C} - T^{R}_{C}T_{C_{j,i+1}}^{C_{j,i}}}^{2}
  + c_{Zi}
  \right) 
  + c_{Ji} \right)
\label{eq:joint_minimization_eq}
\end{split}
\end{equation}

In the following, Joint-MEMROC will be used to denote the calibration process with joint optimization. By exploiting the mutual constraints between camera poses introduced by overlapping fields of view, our extensive experiments will demonstrate that Joint-MEMROC significantly enhances calibration accuracy. This improvement comes from incorporating additional geometric information derived from overlapping areas, enriching the calibration process by providing a more detailed understanding of the sensor configuration.

\section{DATASET}
\label{sec:dataset}
\begin{figure*}[t]
  \centering
  \includegraphics[width=1\linewidth]{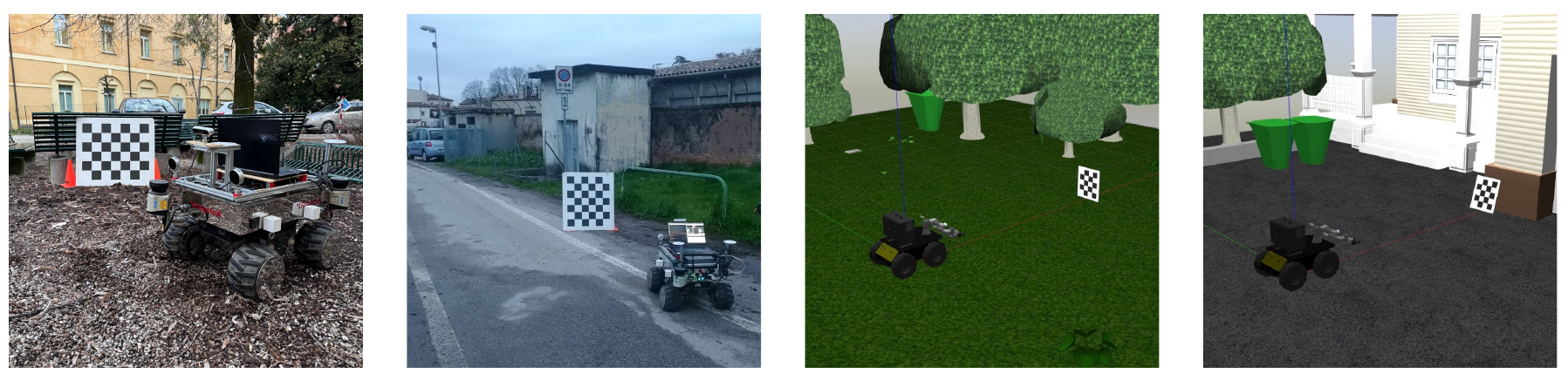}
  \caption{Set of 4 images representing 4 different scenarios where the dataset was collected. The first two on the left show two different real-world scenarios, the other two simulated scenarios.}
  \label{fig:dataset}
\end{figure*}

We collected a dataset\footnote{\url{https://github.com/davidea97/MEMROC.git}} comprising 3000 synthetic images and 3000 real images, to comprehensively validate the proposed method for calibrating a multi-camera system fixed on a mobile robot. 
The synthetic images were generated using the Gazebo simulation toolbox, wherein we utilized a Husky Unmanned Ground Vehicle (UGV) from Clearpath Robotics\footnote{\url{https://clearpathrobotics.com/}} equipped with three simulated Intel RealSense Depth D455 sensors. We conducted 10 independent image acquisition runs across diverse environments, each featuring distinct background and floor textures, as shown in Figure~\ref{fig:dataset}. For each run, the mobile robot was moved in 100 different poses, at varying distances and orientations in front of a 5$\times$4 checkerboard pattern with 10\,cm squares. 
During each pose, all three sensors simultaneously captured images, their corresponding 3D point clouds, and the robot's odometry (i.e., its position and orientation relative to the starting position).
The synthetic dataset includes ground truth information consisting of:
\begin{itemize}
    \item The geometric transformations of each sensor relative to the mobile robot's reference frame;
    \item The relative rototranslations among the various sensors of the multi-camera system.
\end{itemize}

The real part of the dataset consists of images captured by the mobile robot's three onboard cameras, all front-facing, from Robotnik\footnote{\url{https://robotnik.eu/}}. The robot is equipped with two Intel RealSense LiDAR L515 sensors and one Intel RealSense Depth D455 sensor, each designed with overlapping fields of view (FoV).
While D455 cameras can be used both outdoor and indoor, L515 cameras are designed especially for indoor applications and provides very low-quality point cloud when used outside; in the real dataset we consider both sensor to prove the robustness and generality of our approach.

The real-world data consists of images captured in 10 diverse scenarios: 6 indoor and 4 outdoor environments. We collected the data by driving the mobile robot in 100 poses at various distances and orientations with respect to a 6$\times$5 checkerboard with 11\,cm squares. During data acquisition, all three sensors captured images simultaneously together with their corresponding 3D point clouds and the robot's odometry. 

The indoor data encompasses two distinct environments showcasing variations in ground surfaces, background textures, and the inclination and positioning of the calibration pattern. This diversity mainly aims to challenge the calibration methods with different visual features.
The outdoor data focuses on capturing images with diverse backgrounds and varying checkerboard inclinations. This data is particularly valuable for testing the robustness of calibration methods against potential lighting variations and when the robot's odometry becomes very noisy due to the inherent drift experienced on uneven terrains.

Crucially, for both the synthetic and real datasets, we maintained the relative positions of the cameras throughout the entire data collection process. This allows us to evaluate the consistency of calibration methods across multiple runs.

This dataset is valuable for comparing various multi-camera calibration methods. The availability of ground truth information enables to assess how well they perform with or without the calibration patterns, their accuracy with varying image quantities and their effectiveness in different scenarios.

Furthermore, in our dataset a checkerboard pattern is visible in all images: it can be used not only for calibration but also as ground truth for the evaluation of visual odometry methods~\cite{zheng2019visual, wei2022gclo}. To the best of our knowledge, no other publicly available dataset offers this combination of features, making it a beneficial resource for the research community.

\section{EXPERIMENTS AND RESULTS}
\label{sec:experiments}
We conducted several experiments on the collected data to demonstrate the effectiveness and robustness of our calibration method in practice. First, we validated our method using synthetic data, whose ground truth parameters are known (Section \ref{subsec:synthetic}). Next, we assessed the calibration accuracy in the real-world environments (Section \ref{subsec:real}).
We considered both datasets for comparing the proposed methods against the state-of-the-art calibration approaches, namely the motion-based method proposed by Zuñiga et al.~\cite{zuniga2019automatic} and the learning-based method proposed by Yan et al.~\cite{yan2023sensorx2car}. They both  calibrate sensors with respect to the vehicle reference frame, but the latter calibration method only provides rotation calibration parameters, therefore only rotation errors will be reported in the resulting tables. To compare calibration performance, we also assessed the accuracy of estimating the transformations between the various sensors in the multi-camera system. To establish a reliable baseline for our measurements in real-world calibration parameter experiments, we utilized the stereo calibration method developed by Li et al.~\cite{li2013multiple} as the ground truth reference. This method will also be used as a comparison in the simulated dataset to show its correctness and accuracy.

\subsection{Experiments on synthetic dataset}
\label{subsec:synthetic}
The primary goal of using synthetic data is to evaluate the correctness and accuracy of our calibration method by directly comparing it to the ground truth. This allows us to compute the translation error across the three axes and the rotation error around the three rotation axes, expressed in terms of roll, pitch, and yaw. This metric was used to evaluate first the precision of the estimated camera-to-robot transformation and then we assess the consistency of the transformations among cameras by evaluating the camera-to-camera rototranslation error. This helps us identify any potential cumulative errors that might arise during the calibration process.

Tables~\ref{tab:h2e_synth_results} and \ref{tab:c2c_synth_results} report the average error for each calibration parameter of camera-to-robot and camera-to-camera transformations, respectively. These errors represent the mean values calculated across all 10 synthetic datasets.

\begin{table}[h]
\caption{Average error in Camera-to-Robot transformation}
\label{tab:h2e_synth_results}
\centering
\resizebox{\columnwidth}{!}{
\begin{tabular}{|c|ccc|ccc|}
\hline
\multirow{2}{*}{Method} & \multicolumn{3}{c|}{Translation parameters} & \multicolumn{3}{c|}{Rotation parameters} \\
\cline{2-7}
                        & $t_{x}$ [cm] & $t_{y}$ [cm] & $t_{z}$ [cm] & $r_{x}$ [deg] & $r_{y}$ [deg] & $r_{z}$ [deg] \\
\hline
MEMROC & \pmb{$0.61$} & $0.25$ & $0.15$ & \pmb{$0.08$} & \pmb{$0.13$} & $0.16$ \\
Joint-MEMROC & $0.66$ & \pmb{$0.16$} & \pmb{$0.14$} & $0.14$ & $0.14$ & \pmb{$0.15$} \\
\hline
RobotAutocalib~\cite{zuniga2019automatic} & $0.92$ & $0.79$ & $0.92$ & $1.87$ & $1.32$ & $1.84$ \\
SensorX2Car~\cite{yan2023sensorx2car} & $-$ & $-$ & $-$ & $0.26$ & $0.82$ & $1.22$ \\
\hline
\end{tabular}}
\vspace{-0.2cm}
\end{table}

\begin{table}[h]
\caption{Average error in Camera-to-Camera transformation}
\label{tab:c2c_synth_results}
\centering
\resizebox{\columnwidth}{!}{
\begin{tabular}{|c||c|c|c||c|c|c|}
\hline
\multirow{2}{*}{Method} & \multicolumn{3}{c||}{Translation parameters} & \multicolumn{3}{c|}{Rotation parameters} \\
\cline{2-7}
                        & $t_{x}$ [cm] & $t_{y}$ [cm] & $t_{z}$ [cm] & $r_{x}$ [deg] & $r_{y}$ [deg] & $r_{z}$ [deg] \\
\hline
MEMROC & $0.42$ & \pmb{$0.23$} & $0.87$ & \pmb{$0.11$} & {$0.18$} & $0.15$ 
\\Joint-MEMROC & \pmb{$0.20$} & $0.25$ & $0.78$& $0.22$ & $0.23$ & \pmb{$0.14$} \\
\hline
RobotAutocalib~\cite{zuniga2019automatic} & $2.58$ & $4.99$ & $3.72$ & $2.52$ & $3.09$ & $1.68$ \\
SensorX2Car~\cite{yan2023sensorx2car} & $-$ & $-$ & $-$ & $0.44$ & $0.21$ & $2.40$ \\
Li~\cite{li2013multiple} & $0.26$ & $0.27$ & \pmb{$0.34$} & $0.12$ & \pmb{$0.14$} & $0.17$ \\

\hline
\end{tabular}}
\end{table}

Our method achieves higher calibration accuracy for both translation and rotation errors compared to existing state-of-the-art approaches. Notably, RobotAutocalib  struggles with camera-to-camera transformations, possibly due to its disregard for relative constraints existing among cameras. Conversely, the Li stereo calibration method performs well in this specific scenario. Yan's method, despite being learning-based, demonstrates strong accuracy and generalizability across various scenarios, particularly for rotation parameters. 
\subsubsection*{Analysis on image quantity}
the synthetic dataset enables us to evaluate the impact of the number of images on the effectiveness of our multi-sensor calibration method.
\begin{figure}[h!]
  \centering
  \includegraphics[width=1\linewidth]{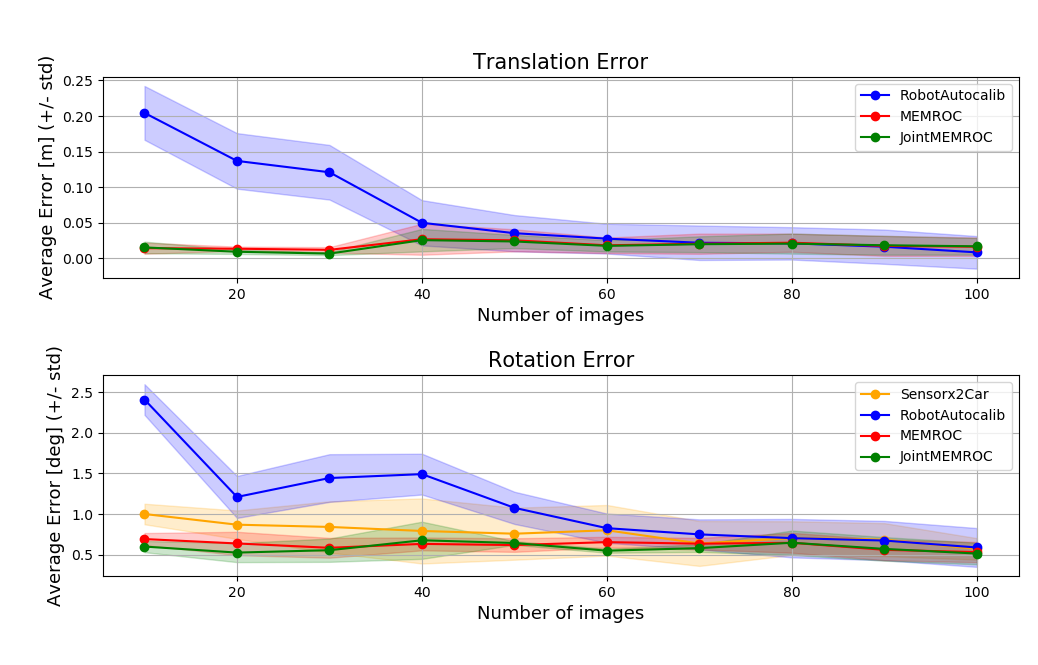}
\caption{Impact of image quantity on calibration methods, where the lines represent the average errors and the shaded areas indicate the standard deviation, reflecting the variability and consistency of each method.}
  \label{fig:h2e_image_plot}
\end{figure}

Figure~\ref{fig:h2e_image_plot} illustrates the substantial impact of image quantity on the effectiveness of calibration methods. From this point of view MEMROC shows remarkable robustness and accuracy even with minimal data available for the calibration, outperforming the motion-based method by Zuñiga's algorithm and the learning-based Yan's approach. 
\subsubsection*{Analysis with noisy odometry}
\begin{figure}[h!]
  \centering
  \includegraphics[width=1\linewidth]{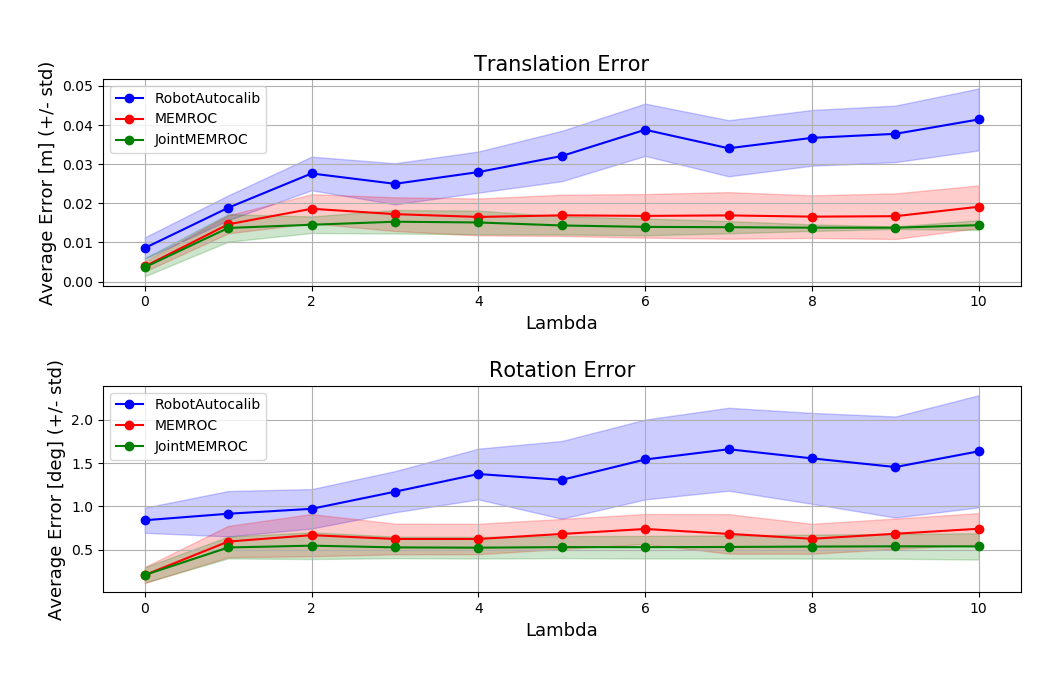}
\caption{Evaluation of motion-based calibration method with incremental noise in measurements provided by odometry, with $\lambda\in[0,10]$. The lines represent the average errors and the shaded areas indicate the standard deviation.}
  \label{fig:h2e_noisy_plot}
  \vspace{-0.15cm}
\end{figure}

the synthetic dataset enabled us to test the robustness and efficiency of our method in challenging scenarios, specifically where odometry measurements are compromised by noise. To thoroughly evaluate robustness in such conditions, we introduced Gaussian noise  $N(0, \sigma_{t})$ for translation, with $\sigma_{t} = \lambda\cdot 0.2$\,cm and Gaussian noise $N(0, \sigma_{r})$ for rotation, with $\sigma_{r} = \lambda\cdot 0.01$\,rad where $\lambda\in[0,10]$. Noise primarily affects odometry measurements in the x and y axes for translation, and in the z-axis for rotation, reflecting the 2D nature of the motion estimates. Figure~\ref{fig:h2e_noisy_plot} illustrates how increasing noise levels in odometry measurements affect the average translation and rotation errors in camera-to-robot transformations. As a result, the comparison in this figure is limited to motion-based calibration methods, since they rely on odometry data.

Figure~\ref{fig:h2e_noisy_plot} demonstrates that the proposed motion-based method exhibits significantly higher accuracy and robustness compared to the RobotAutocalib method under noisy conditions. Notably, our method achieves translation errors consistently below 2\,cm and rotation errors below 1$^\circ$. This highlights its suitability for real-world scenarios with noisy measurements.
Our experiments with synthetic data demonstrate the  efficiency of the proposed method in two key ways. First, they enable the calibration of all 6-DoF for both camera-to-robot and camera-to-camera transformations. Second, they achieve high accuracy, surpassing the motion-based method by Zuñiga and the learning-based by Yan. The ability to calibrate with just a few images highlights that the system is particularly suited for real-world scenarios where a fast and user-friendly approach is crucial.

\subsection{Experiments on real-world scenarios}
\label{subsec:real}
\begin{figure}[h!]
  \centering
  \includegraphics[width=1\linewidth]{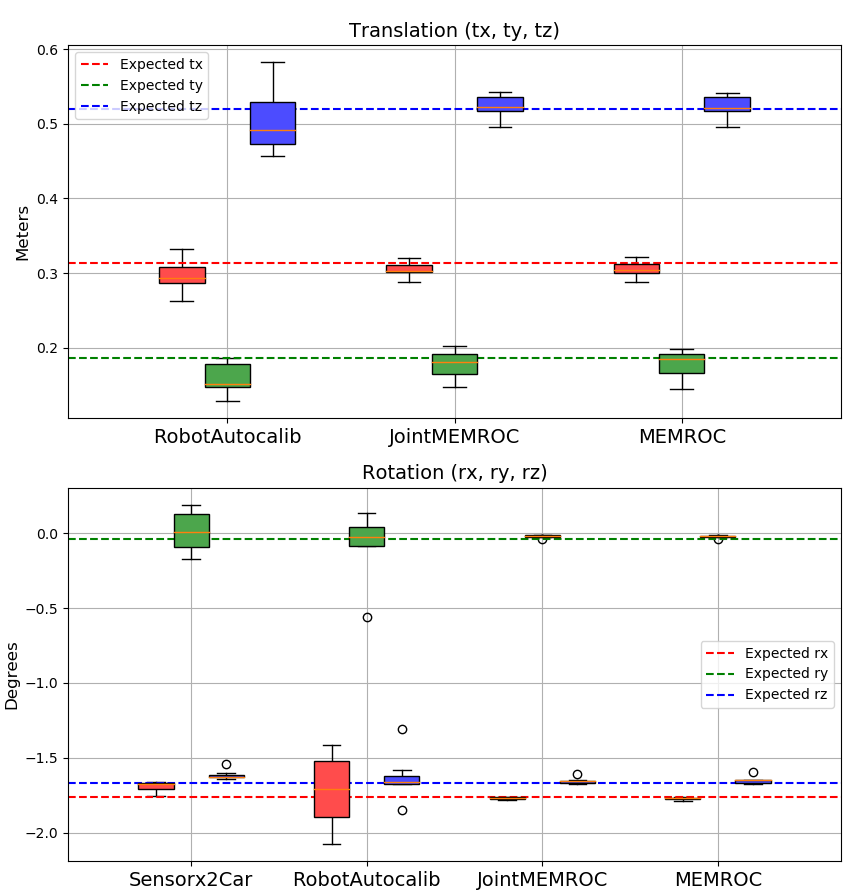}
\caption{Calibration parameters with average and standard deviation of the Intel RealSense LiDAR L515 sensor estimated by different calibration methods. The expected values were measured by using an AprilTag fixed at a precise distance from the mobile robot in front of the sensors.}
  \label{fig:h2e_plot}
  \vspace{-0.2cm}
\end{figure}

To comprehensively assess the effectiveness of our proposed calibration methods, we conducted a series of experiments in real-world indoor and outdoor scenarios. These experiments allowed us to analyze not only the precision of the calibration but also its robustness under non-ideal conditions. This includes situations where odometry measurements suffer from drift
errors. Notably, these real-world tests were crucial for demonstrating the consistency of the results across diverse environments with varying visual-features and floor types, which can significantly impact odometry accuracy.
We evaluated our calibration method against state-of-the-art approaches on 10 real-world datasets, each containing 300 images. Fig.~\ref{fig:h2e_plot} shows for simplicity only the average value and standard deviation obtained by MEMROC and state-of-the-art methods across all runs for each calibration parameter ($t_{x}$, $t_{y}$, $t_{z}$ for translation and $r_{x}$, $r_{y}$, $r_{z}$ for rotation) of the Intel RealSense LiDAR L515 sensor. 

Our experiments demonstrate that MEMROC consistently outperform existing approaches. It achieves higher accuracy on average, and it exhibits greater consistency across multiple runs of the algorithm, resulting in reliable and precise measurements. 

\begin{table}[h]
\caption{Average error in Camera-to-Camera transformation}
\label{tab:c2c_real_results}
\centering
\resizebox{\columnwidth}{!}{
\begin{tabular}{|c||c|c|c||c|c|c|}
\hline
\multirow{2}{*}{Method} & \multicolumn{3}{c||}{Translation parameters} & \multicolumn{3}{c|}{Rotation parameters} \\
\cline{2-7}
                        & $t_{x}$ [cm] & $t_{y}$ [cm] & $t_{z}$ [cm] & $r_{x}$ [deg] & $r_{y}$ [deg] & $r_{z}$ [deg] \\
\hline
MEMROC & 1.97 & 1.15 & 0.34 & 0.79 & {1.02} & {0.72} \\
Joint-MEMROC & \pmb{0.48} & \pmb{0.56} & \pmb{0.33} & \pmb{0.74} & \pmb{0.83} & \pmb{0.66} \\
\hline
RobotAutocalib~\cite{zuniga2019automatic} & 2.41 & 4.42 & 2.43 & 3.83 & 4.82 & 4.22 \\
SensorX2Car~\cite{yan2023sensorx2car} &- & - & - & 2.95 & 1.99 & 2.75 \\
\hline

\hline
\end{tabular}}
\end{table}

Table \ref{tab:c2c_real_results} presents the average errors for relative calibration parameters between the three cameras. Our method achieves superior calibration accuracy compared to both state-of-the-art motion-based and learning-based approaches. It should be emphasized that the MEMROC achieves even higher precision through joint optimization of all sensors, leveraging their overlapping fields of view. This technique is particularly effective for multi-sensor calibration.

\section{CONCLUSIONS}
\label{sec:conclusions}
In this paper we presented MEMROC, a novel method for calibrating multiple sensors with respect to a mobile robot. 
Utilizing robust ground plane detection, MEMROC addresses z-unobservability in planar motion, enabling the estimation of cameras' full 6-DoF pose with just a few images of a simple calibration pattern.
%
Moreover, when sensors have overlapping fields of view, MEMROC exploits this condition optimizing also the geometric constraints existing between sensors. 
A comprehensive analysis of our method against state-of-the-art methods demonstrates that it is highly accurate and more robust than other methods in non-ideal conditions, when the odometry is affected by noise. Moreover, MEMROC proves to be suitable for daily use and for real-world scenarios as it requires only minimal data to achieve higher calibration accuracy.
Currently, we are investigating how to completely remove the checkerboard while maintaining the same performance level. We also plan to extend the technique to sensors like IMU and Laser 2D.

\section*{ACKNOWLEDGMENT}
The research leading to these results has received funding from the European Union's Horizon 2020 research and innovation programme under grant agreement No. 101006732 (DrapeBot).
\vspace{-0.1cm}

\bibliographystyle{IEEEtran}
\bibliography{references.bib}

\end{document}